\documentclass[letter]{IEEEtran}
\usepackage{graphicx}
\usepackage{subfig}
\usepackage{amsmath}
\usepackage{color}
\usepackage{cite}
\usepackage{amsmath,amssymb,amsfonts}
\usepackage{algorithm}
\usepackage{algorithmic}
\usepackage{textcomp}
\usepackage{multirow}
\usepackage{booktabs}
\usepackage{epstopdf}
\usepackage{array}
\usepackage{pgffor}
\usepackage{threeparttable}
\usepackage{multicol}
\usepackage{mathtools}
\usepackage[marginal]{footmisc}
\usepackage{arydshln}

\usepackage[colorlinks,linkcolor=green]{hyperref}

\begin{document}

\title{Random Forest-Based Prediction of Bone Volume Fraction and Fracture Position from S-Parameters}

\author{\IEEEauthorblockN{Jianhe Li, Jinsui Meng, Yida Zhao, Zihe Wang, Liaoran Sun, Tao Shan,~\IEEEmembership{Member,~IEEE}}
	\thanks{This work was supported in part by the National Natural Science Foundation of China under Grant 62401035; in part by the Fundamental Research Funds for the Central Universities; in part by Key Laboratory of Intelligent Systems and Equipment Electromagnetic Environment Effect, Ministry of Industry and Information Technology.(Corresponding author: Tao Shan.)}
	\thanks{Jianhe Li, Tao Shan are with School of Electronics and Information Engineering, Beihang University.
		\\ (e-mail:taoshan@buaa.edu.cn).}}
\maketitle
\begin{abstract}
In this paper, we propose a method for predicting bone volume fraction (BVF) and fracture position by constructing a random forest model based on multichannel S-parameters. A nine-antenna microwave scanning system is designed and fabricated to acquire the multichannel S-parameter data. Bone-mimicking phantoms are developed, and corresponding experiments are conducted to validate the effectiveness of the proposed approach. Both synthetic and experimental results demonstrate the validity of the method.
\end{abstract}
\begin{IEEEkeywords}
	Bone Volume Fraction, Fracture, Random Forest, S parameter
\end{IEEEkeywords}
\section{Introduction}
Osteoporosis and fracture are the two most serious bone health threats facing humanity in the context of global aging. According to World Health Organization (WHO) and Global Burden of Disease (GBD) Study, up to 37 million fragility fractures occur annually in individuals aged over 55 worldwide, and approximately 6.3\% of men over the age of 50 and 21.2\% of women over the same age are affected by osteoporosis range globally \cite{liang_global_2025, wu_global_2021, kanis_reference_2008}. Current medical imaging techniques for bone diagnose such as the
X-ray radiography, computed tomography (CT) and magnetic resonance imaging (MRI), are either radiation-harmful or associated with high cost and bulky equipment, which is not suitable for use in scenarios such as home, outdoor, or surgical settings.

Recent years, microwave sensing (MWS) has been proposed for biomedical imaging and other related fields due to its advantages in non-ionizing radiation and low cost \cite{alkhodari_monitoring_2021}. In MWS, the region of interest (ROI) is typically surrounded by a circular antenna array. By transmitting low-energy microwave signals into the target object and then detecting the scattered fields it generates, the distributions of electrical conductivity and permittivity within the object can be reconstructed. Through subsequent signal processing, imaging information and other relevant metrics of the object are obtained. Researches and applications about MWS include the breast cancer detection \cite{klemm_microwave_2009}, bone lesion detection \cite{alkhodari_monitoring_2021, ruvio_microwave_2016, khalesi_free-space_2020}, and lung cancer diagnose \cite{semenov_microwave_2009}.

Meanwhile, deep learning has a wide applications in microwave sensing and medical diagnose. Artificial neural networks (ANN) can improve the imaging resolution by generating high frequency data from the low frequencies \cite{abir_high_2024}. ANNs like U-Net can also be powerful in the field of medical image segmentation\cite{wang_medical_2022}.

In this work, method integrating microwave sensing with deep learning for the prediction of human bone BVF and fracture position is further explored. We designed a nine-antennas microwave scanning system to probe a cylindrical region. The measured S parameters are processed by Random Forest for BVF and fracture position prediction. To train the Random Forest network, we modeled and simulated the human bone and antenna array. The modeled bones incorporate 10 distinct BVF values and 450 different fracture scenarios, resulting in a total of 4500 cases. The Random Forest network was then trained on datasets of S-parameters, BVF values, and fracture location indicators. Lastly, we fabricated six different bone-mimicking phantoms and the MWS device, conducted physical experiments and evaluated the prediction abilities of the trained network.

\section{The Proposed Method}
\begin{figure*}[h]
	\centering
	\includegraphics[width=0.95\textwidth]{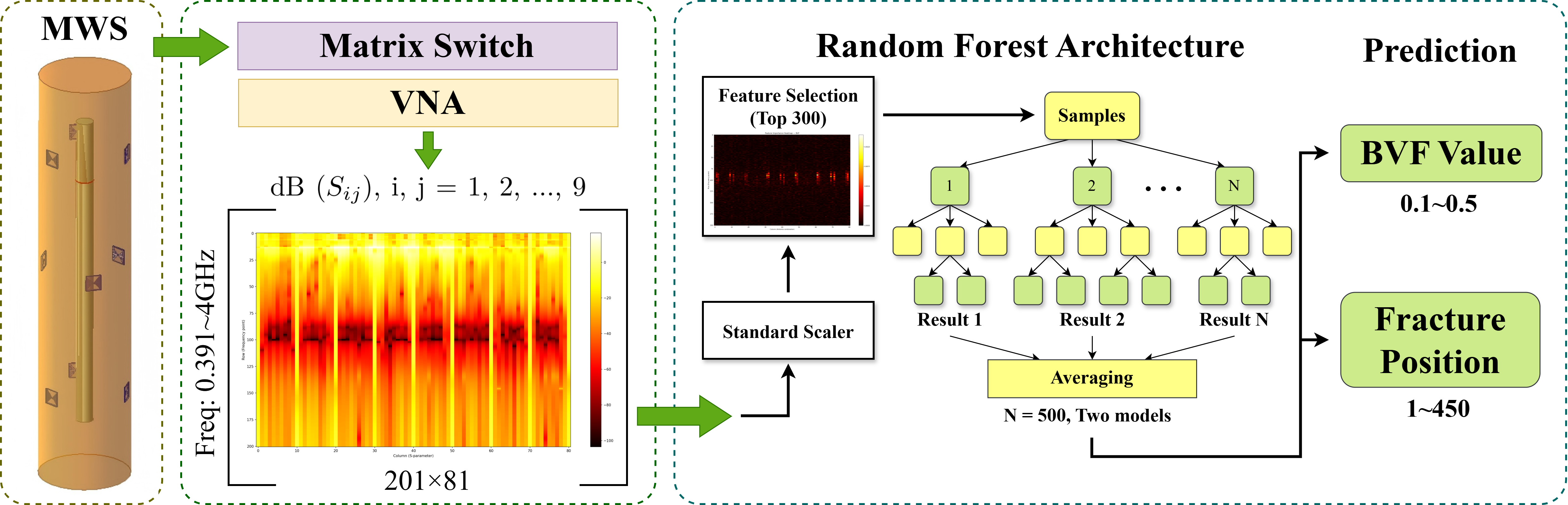}  
	\caption{Overview of the method}
	\label{fig:system overview}
\end{figure*}

The method we proposed for predicting BVF and fracture position using Random Forest consists of two parts: microwave sensing system and Random Forest network. In the first part, the MWS probes the ROI to acquire its dielectric properties. Using the matrix switch and vector network analyzer (VNA), the multichannel S-parameters are obtained. In the second part, the Random Forest network predicts BVF value and fracture position based on processed S-parameters from previous steps. Overview of the proposed method is shown in Fig. \ref{fig:system overview}.

\subsection{Microwave Sensing System}
\begin{figure}[h]
	\centering
	\includegraphics[width=0.3\textwidth]{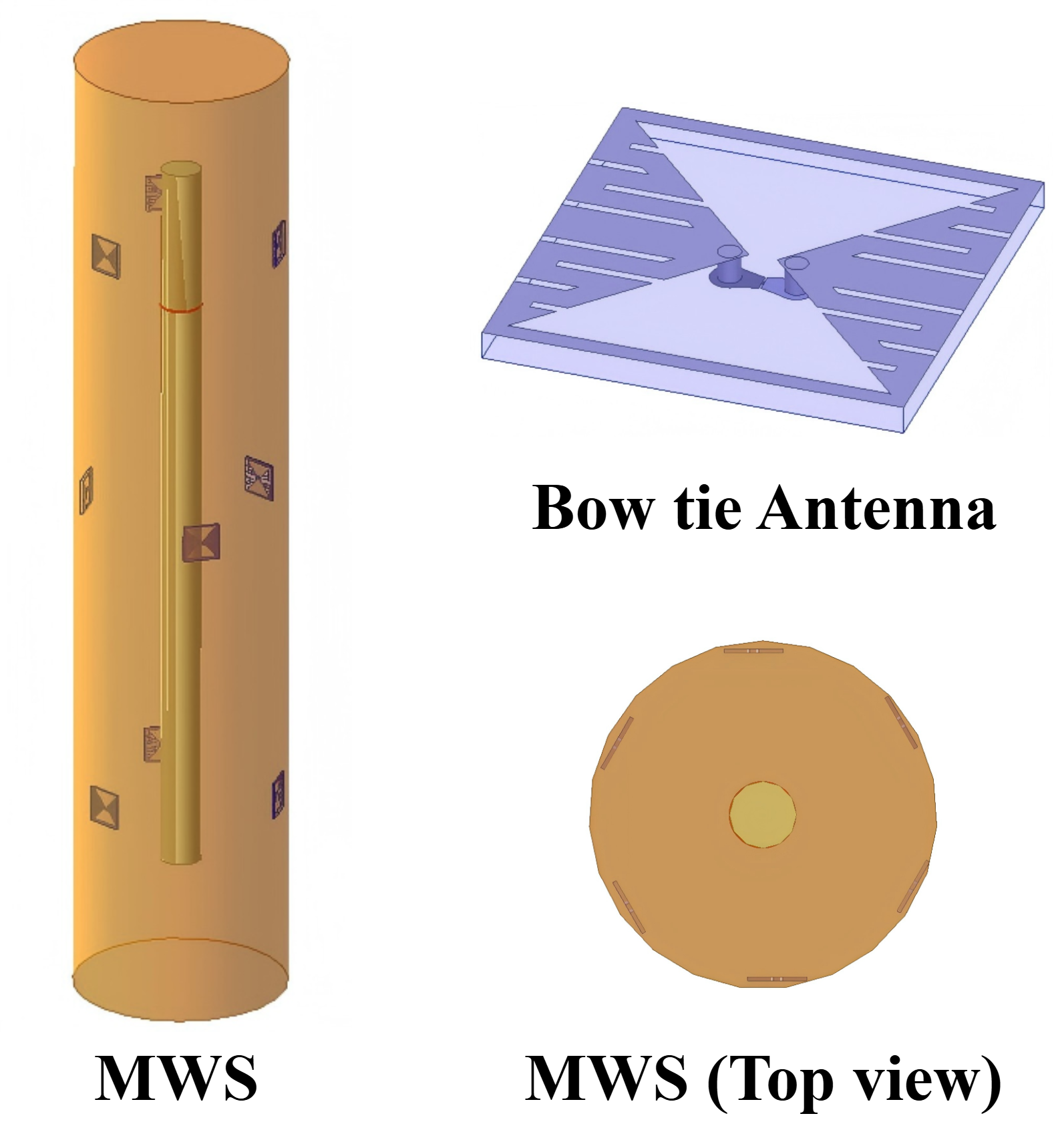} 
	\caption{MWS and Bow Tie Antenna}
	\label{fig:MWS_bowtie_antenna}
\end{figure}

The microwave sensing system is designed as a nine-antenna array with a cylindrical ROI. 
The antennas are placed on three levels at different heights, with three antennas spread evenly around in a circle on each level. 
The antennas on levels next to each other are turned to point in different directions, as shown in Fig. \ref{fig:MWS_bowtie_antenna}. 
The measurement process is designed as follows: A single antenna transmits while the remaining eight receive, measuring the S-parameters between all antenna pairs. 
This process is repeated, cycling the transmitter role through all nine antennas, to complete one full measurement. 
The antenna employs an ultra-wide band (UWB) bow-tie structure on an FR4 substrate with a relative permittivity of 4.4 and a loss tangent of 0.02. 
The  antenna has an overall size of 30 mm × 30 mm × 1.5 mm, covering an operational bandwidth from 391 MHz to 4 GHz, as shown in Fig. \ref{fig:MWS_bowtie_antenna}.

To simplify the model, we modeled the leg and the bone as cylinders, with the bone being placed inside the leg model using a Boolean subtraction operation. 
A cylindrical disc (crack) with a height of 1 mm and a radius equal to that of the bone cylinder is modeled on the leg bone to simulate a fracture. Different fracture locations are simulated by varying the position of this disc within the bone. 
According to work \cite{alkhodari_monitoring_2021} and datasets in \cite{gabriel_dielectric_1996, itis_tissue_v5}, we assigned electrical properties, including conductivity and permittivity, to the leg, bone, and crack models. 
The detailed dimensions and electrical parameters of each model component are provided in Table \ref{tab:Assignment of electrical parameters for each material}. 
Note that \$ef and \$dlt are two variables, changing synchronously to simulate different BVFs \cite{alkhodari_monitoring_2021}. 
Their specific values are listed in Table \ref{tab:efdlt}. 
BVF = 0.5 represents healthy bone, while 0.1 indicates a severe bone mass loss \cite{alkhodari_monitoring_2021}. 
The electrical properties of crack are set equal to leg to simulate fracture cases.

\begin{table*}[t]
	\centering
	\begin{tabular}{ccccccc}
		\toprule
		Component & Height & Radius & Permittivity & Permeability & Conductivity & Dielectric Loss Tangent \\
		\midrule
		leg & 600 mm & 65 mm & 35 & 1 & 0.5 s/m & 0 \\
		crack & 1 mm & 12.5 mm & 35 & 1 & 0.5 s/m & 0 \\
		bone & 450 mm & 12.5 mm & \$ef & 1 & 0.5 s/m & \$dlt  \\
		\bottomrule
	\end{tabular}
	\caption{Assignment of Electrical Parameters for Each Material}
	\label{tab:Assignment of electrical parameters for each material}
\end{table*}

\begin{table}[h]
	\centering
	\begin{tabular}{cccc}
		\toprule
		$\tilde{\epsilon}_{r}$ & \$ef & \$dlt & BVF  \\
		\midrule
		13.000-j3.000 & 13.000 & 0.230 & 0.500 \\
		
		14.111-j3.117 & 14.111 & 0.221 & 0.456  \\
		
		15.222-j3.224 & 15.222 & 0.212 & 0.411   \\
		
		16.333-j3.310 & 16.333 & 0.203 & 0.367 \\
		
		17.444-j3.376 & 17.444 & 0.194 & 0.322  \\
		
		18.556-j3.422 & 18.556 & 0.184 & 0.278  \\
		
		19.667-j3.448 & 19.667 & 0.175 & 0.233 \\
		
		20.778-j3.454 & 20.778 & 0.166 & 0.189 \\
		
		21.889-j3.439 & 21.889 & 0.157 & 0.144  \\
		
		23.000-j3.400 & 23.000 & 0.148 & 0.100 \\
		\bottomrule
	\end{tabular}
	\caption{efdlt and BVF}
	\label{tab:efdlt}
\end{table}
\subsection{Random Forest For Prediction of BVF and Fracture Location}
In the prediction task of this work, the model is required to simultaneously regress two key indicators: the BVF value and the fracture position. This problem constitutes a multi-output regression task, which places high demands on the model's nonlinear modeling capability and generalization ability. 
Random Forest, as a non-parametric method based on ensemble learning, can capture complex nonlinear mappings by integrating multiple decision trees, while also offering good resistance to over-fitting \cite{breiman2001random}.
Consequently, we employ two Random Forest regression models to predict BVF and bone fracture position respectively from S-parameters.
The model for BVF prediction processes raw input S-parameter matrix (201 × 81, flattened into 16,281 features) without standardization or feature selection, employing 500 trees with $max-features=1.0$ (all features considered at each split).
For the model predicting fracture position, S-parameters are standardized via Z-score normalization, and the top 300 features are selected for each target based on   impurity-based importance scores from a baseline Random Forest model. The trees number of 500 and
$max-features='1.0'$ are chosen, as these combinations had the best performance in our hyper-parameters tuning experiment.
Both models are trained on the datasets generated via full-wave simulation, comprising a total of 4,500 samples. Each data sample consists of an input S-parameter matrix of dimensions 201 × 81 and a corresponding two-dimensional output label containing the target BVF value and fracture location. The datasets was split into training and test sets in an 8:2 ratio to ensure the independence and reliability of model evaluation.

\section{Numerical Results}
In this section, the prediction result of BVF values and fracture position are shown.
\subsection{Synthetic Data inversion}
In the simulation, a frequency sweep was configured as a linear scan with 201 points ranging from 0.391 GHz to 4 GHz. A parametric sweep on the electrical properties of the bone was also set up, enabling automated simulation of 10 different BVFs for each fracture location. A total of 450 fracture locations were configured, with ten different bone volume fractions simulated for each location, resulting in 4,500 total simulation cases. The whole process is automated through python scripts.
\par
\begin{figure}[h]
	\centering
	\includegraphics[width=0.45\textwidth]{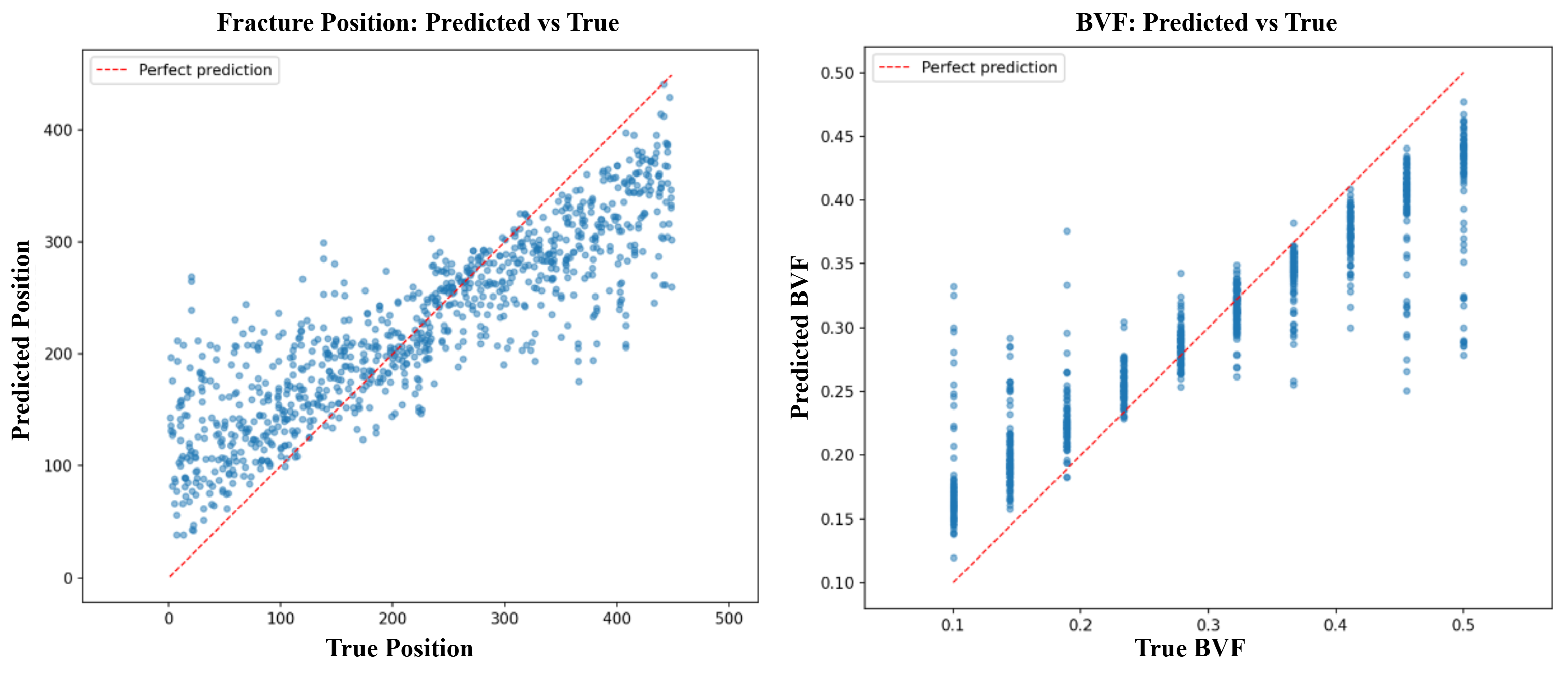}  
	\caption{Prediction Results across Different Fracture Locations (left) and BVFs (right) on Simulated Datasets. Red line represents the perfect prediction. }
	\label{fig:predict sim}
\end{figure}

 
Fig. \ref{fig:predict sim} presents the prediction results of two Random Forest models across different BVFs and fracture locations after training on simulated datasets. It can be observed that most of the predicted fracture locations and BVF values are aligned with the true value with $R^2 = 0.6972, 0.7997$ respectively.
Although some predicted results are not highly precise, the data points still display the correct trend, indicating that models have successfully captured the underlying pattern. Notably, the error is even smaller for positions near the mid-shaft of the bone. Given that bones in real world are often prone to breaking near the middle, these results indicate that the model already possesses a certain level of practical utility.
\subsection{Experimental Results}
A two-layer leg phantom was designed. The outer hollow cylinder models fat and muscle tissue, while the inner solid cylinder models the bone.

Our phantom uses the following raw materials. Distilled water, polyethylene powder, sodium chloride (NaCl), agar, and xanthan gum. To determine the specific composition for a phantom with target permittivity and conductivity, multiple small samples with varying compositions are first prepared. The conductivity and permittivity of each sample are measured, and the sample meeting the requirements is selected as the final formulation. The detailed production process of samples (the same as phantoms) can be found in \cite{islam2018experimental}.

\begin{figure}[h]
	\centering
	\includegraphics[width=0.35\textwidth]{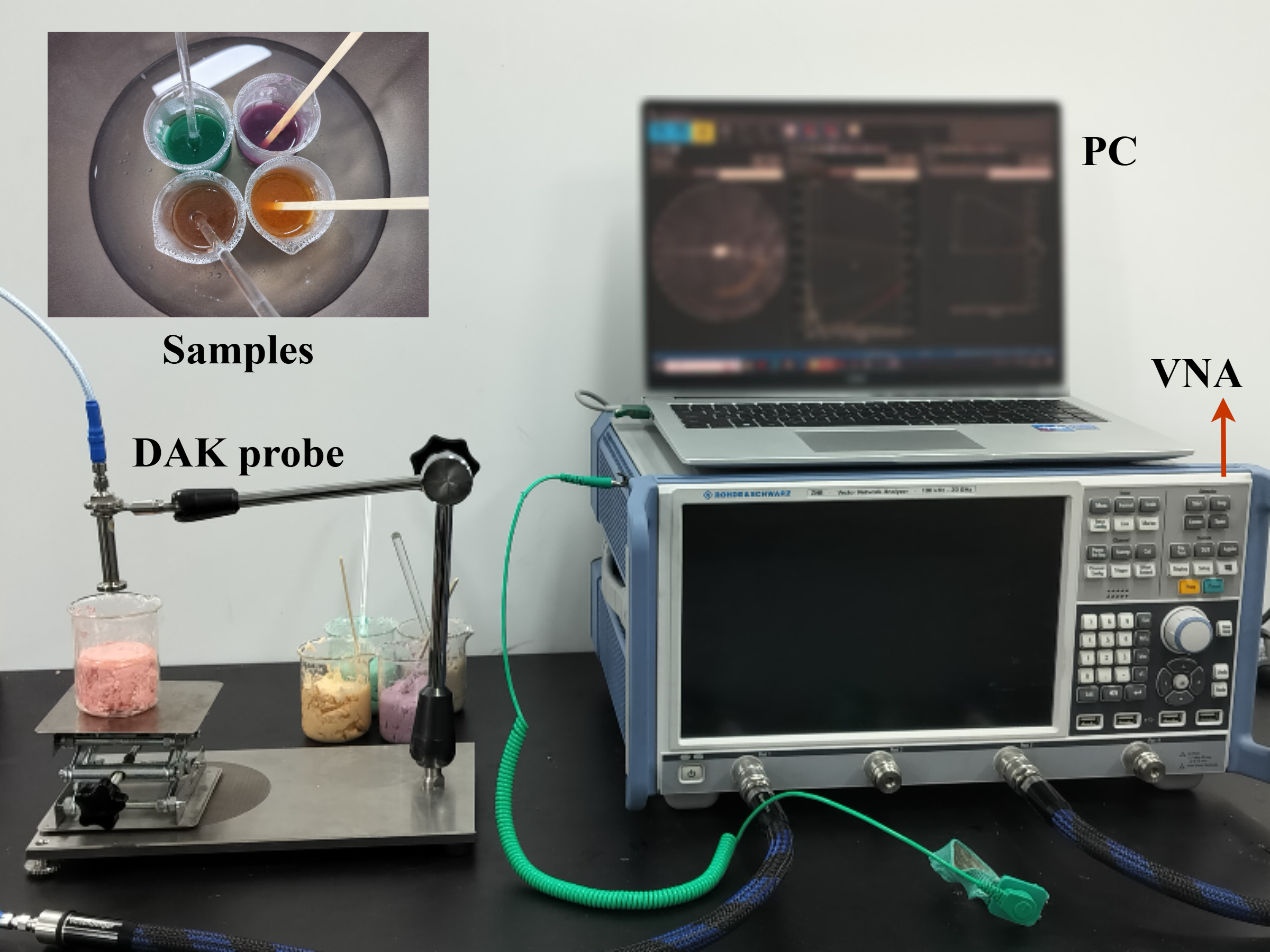}  
	\caption{Setup for permittivity and conductivity measurement (Dielectric probe model:SPEAG DAK SOP)}
	\label{fig:system probe}
\end{figure}

We use a dielectric probe together with a VNA to measure the permittivity and conductivity of the sample. The measurement setup is shown in Fig. \ref{fig:system probe}. Based on the formulations of samples, we got the final composition for phantom production, as listed in table \ref{tab:final composition for phantom production}.

\begin{figure}[h]
	\centering
	\includegraphics[width=0.35\textwidth]{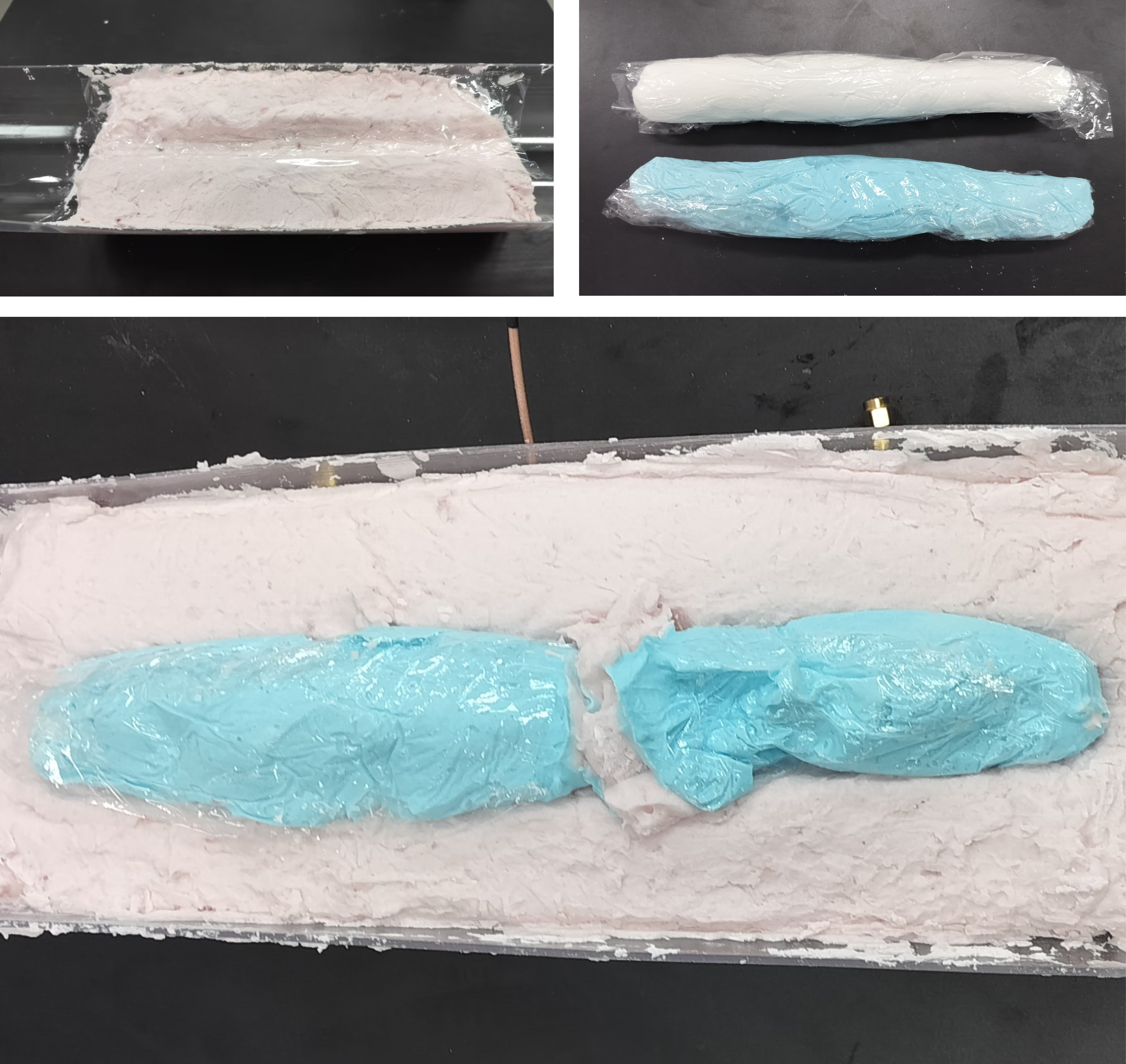}  
	\caption{Bones and the Leg Phantoms}
	\label{fig:fracture bone in leg}
\end{figure}

The bone and leg phantoms were fabricated separately according to their respective formulations. The leg phantom was then divided into two halves and poured into a cylindrical mold. A cavity was created in the center to accommodate the bone phantom.
To simulate a fracture, the bone phantom was truncated and the gap was filled with the leg phantom material, as shown in Fig. \ref{fig:fracture bone in leg}. Table \ref{tab:final composition for phantom production} lists the dielectric parameters of the constructed leg and bone phantoms.

We set up six cases including different fracture locations and permittivity (corresponding to different BVF values). Table \ref{tab:detailed information in each case} lists the detailed information in each case.

\begin{figure}[H]
	\centering
	\includegraphics[width=0.35\textwidth]{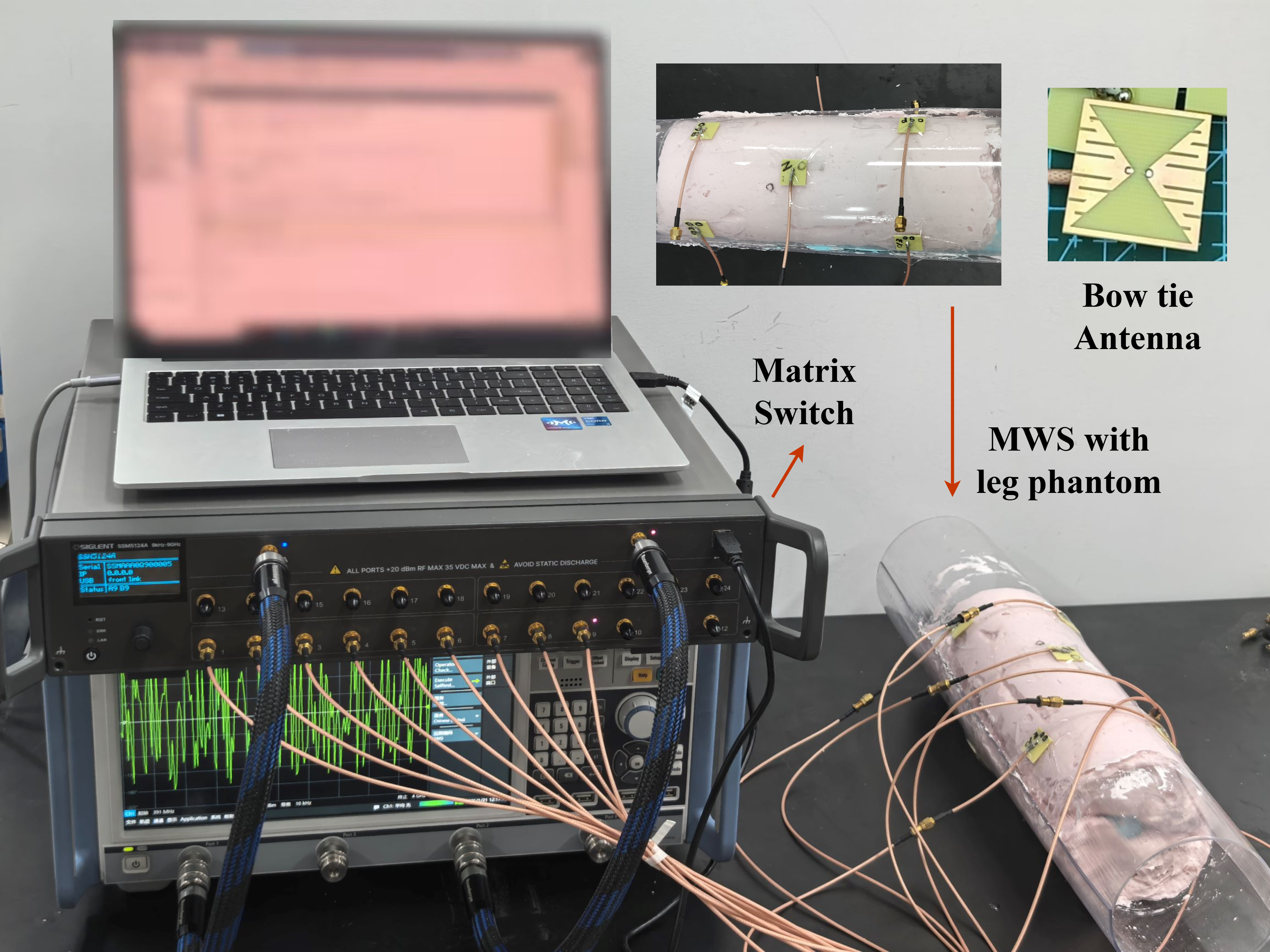}  
	\caption{MWS System}
	\label{fig:system measureS}
\end{figure}

The nine measurement antennas were attached to the leg phantom at positions corresponding to the simulation setup (see Fig. \ref{fig:MWS_bowtie_antenna}, \ref{fig:system measureS}). They were connected to a matrix switch, which was then linked to a VNA for S-parameter measurement. The frequency sweep and other settings were kept identical to those used in the simulation to ensure that the measured S-parameter matrix would be of the same dimensions as $201\times 81$ in simulation setup.

\begin{table}[t]
	\centering
	\begin{tabular}{ccccc}
		\toprule
		Part & Height & Radius & Permittivity & Conductivity \\
		\midrule
		leg & 350 mm & 65 mm & 35 & 0.5 s/m \\
		\midrule
		bone & 300 mm &12.5 mm &\$ef & 0.5 s/m \\
		\bottomrule
	\end{tabular}
	\caption{Settings of Phantom Parameters}
	\label{tab:phantom param settings}
\end{table}

\begin{table}[ht]
	\centering
	\caption{Final Composition for Phantom Production}
	\label{tab:final composition for phantom production}
	\begin{tabular}{l*{3}{c}}
		\toprule
		Part & Bone 1 (white) & Bone 2 (blue) & Leg  \\
		\midrule
		Distilled Water (mL) & 120 &  160 & 2500 \\
		Polyethylene Powder (g) & 120 & 84 &750 \\
		NaCl (g) & 0.3 & 0.4 & 5 \\
		Agar (g) & 5.3 & 7 & 110 \\
		Xanthan Gum (g) & 1.9 &2.6 & 40 \\
		\midrule
		Permittivity & 25 & 32 & 45 \\
		Conductivity (S/m) & 0.6 & 0.6 & 0.7 \\
		\bottomrule
	\end{tabular}
\end{table}

\begin{table}[H]
	\centering
	\caption{Detailed Information in Each Case. ``None'' in column ``Fracture Locations'' means that there's no fracture in this case.}
	\label{tab:detailed information in each case}
	\begin{tabular}{ccc}
		\toprule
		Case Number & Permittivity & Fracture Locations (mm) \\
		\midrule
		1 & 32 & None \\
		2 & 32 & 140  \\
		3 & 32 & 210  \\
		4 & 25 & None \\
		5 & 25 & 140  \\
		6 & 25 & 200  \\
		\bottomrule
	\end{tabular}
\end{table}

Table \ref{tab:rf results physical} shows the prediction results under different physical cases. The sum of BVF values in case 1, 2, 3 is lower than the sum in case 4, 5, 6, indicating that the bone with a lower permittivity has a higher BVF, which is a correct trend. However, the model seems to fail to predict the correct fracture positions.
\begin{table}[H]
	\centering
	\caption{Prediction Results on Real-measured Data}
	\label{tab:rf results physical}
	\begin{tabular}{ccc}
		\toprule
		Case & Prediction & Prediction \\
		Number & BVF & Fracture Locations (mm) \\
		\midrule
		1 & 0.29556 & None \\
		2 & 0.28440 & 234.30  \\
		3 & 0.30018 & 238.68  \\
		4 & 0.28828  & None \\
		5 & 0.29989 & 250.86  \\
		6 & 0.309 & 241.92  \\
		\bottomrule
	\end{tabular}
\end{table}

\section{Conclusion}
This project explores radiation-free prediction of BVF and fracture locations via MWS. We designed a nine-antenna cylindrical detection array, employing two Random Forest models for predicting BVF and fracture location.

System modeling of the antenna array, leg muscle tissue, bone, and simulations of various fracture and BVF scenarios was conducted. The S-parameters between all antenna pairs were obtained from these simulations. Using these S-parameters along with the corresponding ground truths—BVF and fracture locations—we trained the Random Forest models to realize the prediction.

Beyond simulation, a physical measurement platform and two-layered leg phantoms were constructed for experimental validation.The results indicate that the current models fail to effectively predict bone parameters. Potential reasons include:

A)  Inadequate design of the Random Forest models, where the models themselves are ineffective or unsuitable for this specific task.

B)  Over-fitting of the models on the training datasets, hindering their generalization to measured data.

C)  Significant discrepancies between the actual measurement environment and the simulation setup, such as substantial differences in the phantom's dielectric properties, dimensions, or morphology compared to the simulated model.

Future works are needed to address the problems above.

\bibliographystyle{IEEEtran}
\bibliography{IEEEabrv,ref}

\end{document}